\documentclass[conference]{IEEEtran}
\IEEEoverridecommandlockouts
\usepackage{cite}
\usepackage{amsmath,amssymb,amsfonts}
\usepackage{algorithmic}
\usepackage{graphicx}
\usepackage{textcomp}
\usepackage{xcolor}
\def\BibTeX{{\rm B\kern-.05em{\sc i\kern-.025em b}\kern-.08em
    T\kern-.1667em\lower.7ex\hbox{E}\kern-.125emX}}

\usepackage[sort&compress, numbers]{natbib}
\usepackage[T1]{fontenc}
\usepackage[utf8]{inputenc}

\usepackage[colorlinks=true, citecolor=black, linkcolor=black, urlcolor=black]{hyperref}
\usepackage{makecell}
\usepackage{subcaption}
\usepackage{multirow}
\usepackage{stfloats}
\setcitestyle{square}

\begin{document}

\title{YOLO based Ocean Eddy Localization \\with AWS SageMaker\\
\thanks{This work is supported by an Earth Science Informatics Partner (ESIP) lab grant, grant 80NSSC21M0027 from the National Aeronautics and Space Administration (NASA) and grant OAC--1942714 from the National Science Foundation (NSF). Jinbo and Benjamin are partially supported by the NASA Physical Oceanography program and the NASA Physical Oceanography Distributed Data Archive Center (PO.DAAC).}
}
\author{\IEEEauthorblockN{Seraj Al Mahmud Mostafa$^1$, Jinbo Wang$^2$, Benjamin Holt$^2$,  Jianwu Wang$^1$}
\IEEEauthorblockA{\textit{$^1$Department of Information Systems, University of Maryland, Baltimore County,} MD, USA\\
\textit{$^2$Jet Propulsion Laboratory, California Institution of Technology, Pasadena,} CA, USA\\
$^1$\{serajmostafa, jianwu\}@umbc.edu, $^2$\{jinbo.wang, benjamin.m.holt\}@jpl.nasa.gov}
}



\maketitle

\begin{abstract}
Ocean eddies play a significant role both at the sea surface and beneath it, contributing to the sustainability of marine ecosystems and influencing broader oceanic and climatic behaviors. Investigating ocean eddies is essential for monitoring changes in the Earth's oceans and their impact on climate. This study focuses on benchmarking the performance of state-of-the-art YOLO (You Only Look Once) models for locating small-scale (\textless 20km) ocean eddies using satellite remote sensing images. We leverage AWS SageMaker for this evaluation, utilizing both single and multi-GPU configurations to explore the feasibility and efficiency of deploying AI applications in cloud-based environments. This research not only assesses the effectiveness of SageMaker in handling complex Earth science data but also provides insights into deployment challenges, resource management for large-scale data, and the overall user experience. The findings highlight the strengths and limitations of using SageMaker for remote sensing applications and suggest potential future research directions. Our code is open-sourced at \url{https://shorturl.at/hcjmq}. 
\end{abstract}

\begin{IEEEkeywords}
Cloud Services, SageMaker, S3, Ocean Eddy, Localization, Detection, YOLO, YOLOv5, YOLOv8, YOLOv9.
\end{IEEEkeywords}

\section{Introduction}
\label{sec:intro}

Object localization in satellite imagery, particularly for detecting ocean eddies, is crucial for Earth observation due to its impact on understanding marine ecosystems, biodiversity, and climate dynamics. Accurately identifying and tracking these swirling vortices in satellite images enables researchers to monitor sea surface dynamics, including water mass movement, currents, and temperature gradients. Such observations facilitate insights into nutrient cycling and ecosystem behavior, essential for marine conservation efforts.

Despite the advances in technology, processing and analyzing large volumes of satellite imagery for ocean eddy detection poses significant challenges. Solutions that incorporate cloud computing have emerged to address the limitations of traditional, localized computational approaches. Cloud platforms offer scalability, efficiency, and accessibility that are vital for handling extensive Earth observation data. However, the deployment of machine learning applications in the cloud environment is not without its challenges, such as interoperability \cite{giove2013approach}, quality of service \cite{franceschelli2013space4cloud}, and maintaining accuracy \cite{halpern2019one}, responsiveness \cite{ZTunet, AzimFlood, OmarYolov5, islam2021interpreting}, and cost-effectiveness \cite{ardagna2014multi, hussein2019study}. Scalability for large-scale Machine Learning as a Service (MLaaS) \cite{liberty2020elastic} or infrastructure-as-a code \cite{grzegorowski2021cost} remain as notable issues, compounded by challenges specific to climate change applications \cite{wang2021reproducible, yang2017big}.

To address these challenges, our work focuses on deploying and benchmarking State-of-the-art object detection models for ocean eddy localization using Amazon SageMaker, a robust platform offering end-to-end solutions for machine learning model development and deployment. SageMaker's capabilities include integrated development environments (IDEs), built-in algorithms, support for custom solutions, and efficient data management through Amazon S3.

\begin{figure}
    \begin{center}
        \includegraphics[width=.975\linewidth]{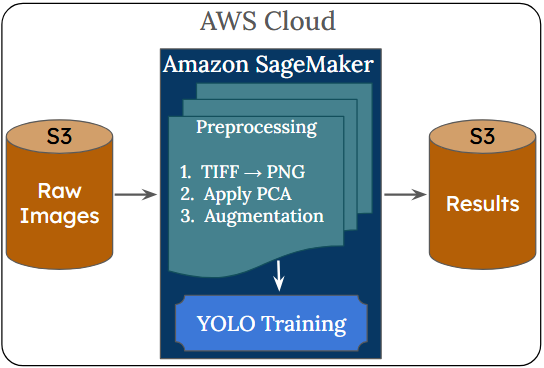}
        \caption{Pipeline for Localizing Ocean Eddies using various YOLO models with SageMaker.}
        \label{fig:arch}
    \end{center}
    \vspace{-2em}
\end{figure}

In this study, we leverage Amazon SageMaker for deploying YOLO models in both single-GPU and multi-GPU configurations to conduct an extensive benchmarking analysis of YOLOv5, YOLOv8, and YOLOv9 for ocean eddy detection. The primary goals of this work are to ensure ease of model deployment, provide access to data, code, pre-built models, and libraries within a cloud-based framework, and analyze their adaptability for real-world applications in oceanography. Our deployment architecture, depicted in Figure~\ref{fig:arch}, outlines the processes of data handling, model training, and result storage within the cloud. Users can construct machine learning pipelines, upload annotated datasets, and execute training processes in SageMaker, with results stored in S3 for efficient data exchange and scalability. The main experiments conducted in this work are highlighted as follows.

\textbf{\textit{I) Data Annotation and Model Training:}} SageMaker's Ground Truth tool was employed for data annotation, focusing on bounding box labeling within images. This tool facilitates accurate preparation of training data for model development. Additionally, the platform's built-in and custom algorithm support allows for flexible model training tailored to specific data types and objectives.

\textbf{\textit{II) Ocean Eddy Detection:}} SageMaker notebooks, akin to Jupyter notebooks, are employed for coding and data analysis. These notebooks interface seamlessly with S3 for data loading and storage. For this work, we deployed YOLO models to perform object localization on satellite imagery, optimizing configurations for both single-GPU and multi-GPU setups.

\textbf{\textit{III) Model Benchmarking:}} We conducted a comparative benchmarking of YOLOv5, YOLOv8, and YOLOv9 to evaluate their performance in detecting ocean eddies. Our analysis includes a comparative study of YOLO metrics such as precision, recall, and mAP scores, as well as hardware-specific metrics, including Giga Floating Point Operations per Second (GFLOPs) and the number of parameters.

This structured approach provides valuable insights into the application of cloud-based deep learning models for large-scale Earth observation data, highlighting their potential to enhance the understanding and monitoring of marine dynamics. Challenges encountered during data annotation and preprocessing are outlined, along with an assessment of the feasibility of deploying these models for real-world oceanographic applications. The experimental methodology, dataset preparation, and implementation details are further discussed in subsequent sections.

The rest of the paper is summarized as follows. We introduce the actual data, the main object of interest, the cloud platform, and the models to be used in Section \ref{background}. Data preprocessing techniques are described in Section \ref{datapreproc}. Section \ref{methods} elaborates on the experimental process in detail. In Section \ref{results}, we discuss the results and benchmark the models used in the experiments, followed by discussions in Section \ref{disc}. Finally, we conclude this report in Section \ref{conc}.

\section{Background}
\label{background}
This section provides essential context for understanding the key elements and terms referenced throughout this study. We provide background information related to the data, the main object of study, known as ocean eddies, and the state-of-the-art model YOLO, which we have explored.

\subsection{SAR Data} 
Synthetic Aperture Radar (SAR) involves active remote sensing, where energy emitted by the sensor is reflected back from the Earth's surface and recorded. Unlike optical imagery, interpreting SAR data requires unique methods as it is influenced by surface properties like structure and moisture. The spatial resolution depends on the ratio of the sensor wavelength to the antenna length, with longer antennas generally producing higher resolution. In this study, SAR data from the European Space Agency's Sentinel-1 mission \cite{ws3} was utilized, providing high-resolution imagery that aids in identifying various geophysical phenomena.

\subsection{Ocean Eddies}
Oceanic circulation is characterized by turbulent structures, including eddies, which are circular water currents. Eddies can range in size from a few kilometers to over 300 kilometers. Mesoscale eddies ($\sim$100-300 km) play a significant role in horizontal transport of water, heat, and tracers, whereas smaller-scale eddies (less than 50 km) are crucial for vertical mixing and interaction with other climate system elements \cite{CHELTON2011167}. Due to spatial limitations of satellite altimetry, small-scale eddies are challenging to study globally. SAR imagery provides an opportunity for such analysis by revealing these eddies at kilometer-level resolution, although the systematic study is hindered by the absence of labeled images. The integration of deep learning methods and scalable cloud-based infrastructure could overcome these barriers, enabling comprehensive surveys.

\subsection{SageMaker}
Amazon SageMaker is a robust cloud service for developing and deploying machine learning models \cite{ws4}. It streamlines workflows by integrating Jupyter notebooks and eliminating server management. SageMaker supports distributed training, GPU utilization, and efficient model deployment, making it a suitable choice for handling large-scale datasets. By leveraging high-performance GPUs such as NVIDIA A100, SageMaker accelerates both training and inference, facilitating rapid prototyping and deployment. This flexibility allows researchers to balance cost-effectiveness and performance, aligning with concerns about achieving optimal quality within budget constraints.

\subsection{YOLO Models} 
The YOLO (You Only Look Once) series, introduced by Redmon et al. \cite{redmon2016you}, revolutionized real-time object detection by predicting bounding boxes and class labels in a single pass. YOLO models approach object detection as a regression task, incorporating CNN-based feature extraction and detection heads for efficient performance. Advancements from YOLOv5 to YOLOv9 have introduced various enhancements in backbone architectures, anchor-based and anchor-free methods, and computational optimizations, improving accuracy and inference speed \cite{selcuk2023comparison}.

\subsubsection{YOLOv5} Developed by Ultralytics, YOLOv5 refines object detection through an anchor-based approach and a Feature Pyramid Network (FPN) backbone \cite{glenn_jocher_2020_4154370}. The FPN is a multi-scale feature extractor designed to enhance the detection of objects at different scales by combining high-resolution features from earlier layers with semantically stronger, lower-resolution features from later layers in the network \cite{lin2017feature}. This hierarchical representation improves the model's ability to identify both small and large objects within an image. YOLOv5 also integrates multi-scale feature extraction and non-maximum suppression to output bounding boxes with objectness scores and class probabilities efficiently.

\subsubsection{YOLOv8} YOLOv8, introduced anchor-free detection, which streamlines model training and reduces the overall parameter count. Unlike traditional anchor-based methods (YOLOv5 and earlier versions) that rely on predefined anchor boxes to predict object locations and sizes, the anchor-free approach allows YOLOv8 to detect objects by directly predicting key points or centers of objects, along with their size and class \cite{farooq2024improved}. This shift eliminates the complexity of anchor box matching and leads to more efficient and flexible object detection. YOLOv8 also incorporates optimized module design, including modified kernel sizes and direct feature attachments in the neck, further enhancing the model's post-processing speed and overall performance \cite{yolov8_ultralytics}.

\subsubsection{YOLOv9} YOLOv9 further improves object detection with the Programmable Gradient Information (PGI) and Gradient Enhanced Learning Architecture Network (GELAN) architectures. PGI optimizes the gradient flow through the network, improving how information is propagated during training. By enhancing the gradient's ability to capture and update model parameters more effectively, PGI prevents issues like gradient vanishing or explosion, which can slow down or destabilize training. On the other hand, GELAN focuses on improving the learning process by enhancing how the model learns complex patterns through a more efficient gradient enhancement mechanism. It supports device-specific block selection, enabling the architecture to be more adaptable and suited for different hardware setups. These innovations help reduce information loss, maintain high detection accuracy, and ensure faster real-time performance across various environments \cite{wang2024yolov9}.

\begin{figure}[t!]
    \begin{center}
        \includegraphics[width=0.975\linewidth, height=4.5cm]{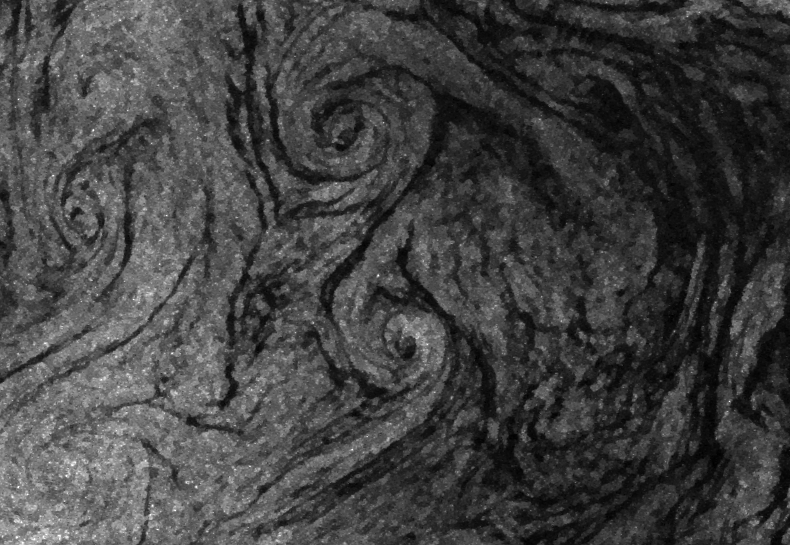}
        \caption{An example of multiple Eddy containing SAR image converted using GDAL library.}
        \label{fig:oe-vis}
    \end{center}
    \vspace{-2em}
\end{figure}

\section{Data Preprocessing} 
\label{datapreproc}
We preprocessed our data in several steps, including initial preprocessing, application of Principal Component Analysis (PCA), and image augmentation. Below, we describe each of these processes in detail.
\subsection{Image Conversion}
The SAR data utilized in this study comprises satellite images stored as \texttt{TIFF} files, each approximately 40 MB in size. To enhance visualization and improve image clarity, we employed \texttt{Quantum Geographic
Information System (QGIS)} software \cite{qgis}, which allowed us to convert the original \texttt{TIFF} images into more manageable form such as \texttt{unsigned integers (UInt8) PNG} format. This step was facilitated by \texttt{Geospatial Data Abstraction Library (GDAL)} \cite{ws14}, a comprehensive library for geospatial data handling that provides seamless support for reading and converting various geospatial formats. By integrating \texttt{GDAL} with Python, we efficiently converted all SAR images into PNG format to ensure compatibility with SageMaker, our chosen cloud-based machine learning platform. The benefits of using \texttt{GDAL} include its compatibility with SageMaker's interface, particularly in the Ground Truth panel, and its ability to reduce image size while preserving quality. An example of a processed image is shown in Figure \ref{fig:oe-vis}. This preprocessing approach aligns with techniques outlined in \cite{mostafa2023cnn}.

\subsection{PCA based Image Restoration}
\begin{figure}[h]
    \begin{center}
        \includegraphics[width=\linewidth]{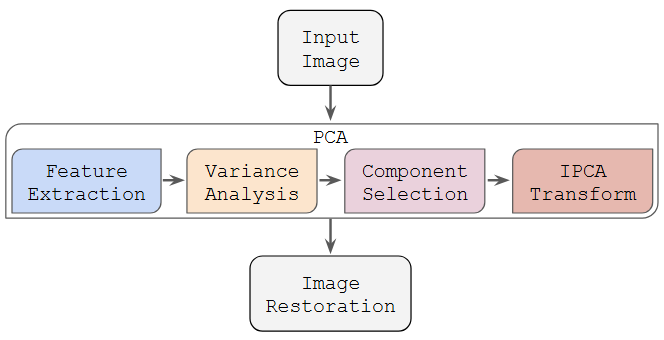}
        \caption{Image restoration process using PCA.}
        \label{fig:pca}
    \end{center}
    \vspace{-2em}
\end{figure}

To address the prolonged training times caused by the increased size and complexity of this expanded dataset, we implemented PCA for image restoration by unwanted feature reduction. The reconstruction process follows several steps, as illustrated in Figure~\ref{fig:pca}. PCA is first fitted to the input image matrix to compute the principal components and their corresponding eigenvalues, followed by calculating the cumulative variance ratio to understand feature importance. The optimal number of components ($k$) is then determined based on the desired variance retention threshold, after which Incremental PCA (IPCA) is applied using the selected $k$ components to project the image data onto a lower-dimensional space. Finally, inverse transformation reconstructs the image from its compressed representation, effectively reducing dimensionality while preserving significant features, with reconstruction quality directly related to retained components. Inspired by the approach in \cite{mostafa2023cnn}, this process effectively streamlined computations and reduced processing time. 

\subsection{Augmentation and Annotation}
We started with 100 image patches, each containing one or more ocean eddies. To augment the dataset, we applied rotations of \(90^\circ\), \(180^\circ\), and \(270^\circ\) to these images, increasing the total number of images to 400. This augmentation resulted in a higher number of eddy occurrences across the dataset, providing more varied examples for training the models and improving their performance.
For annotating images, we used \texttt{LabelImg} software to generate text files in a format suitable for YOLO models \cite{labelimg}. Each text file contains entries specifying the class label of the object, along with bounding box coordinates, which are defined by the top-left \texttt{(x\_min, y\_min)} and bottom-right \texttt{(x\_max, y\_max)} corners of the bounding box. This method enables the efficient labeling of multiple objects within a single image, facilitating the training of YOLO-based object detection models.
For model training, we divided the dataset into 75\% for training, 20\% for validation, and 5\% for testing, ensuring that the augmented images were evenly distributed across all sets for consistency.

\section{Experimental Setup for SageMaker}
\label{methods}
This section details the process of preparing SageMaker for deploying various YOLO models in the cloud environment to benchmark performance on ocean eddy data.

\subsection{Preliminary Steps}
Deploying models in SageMaker involves several key steps. First, users must create or select an existing IAM (Identity and Access Management) role to grant the necessary permissions for instance creation and script execution. Next, users choose an instance with the desired operating system and environment settings (e.g., deep learning) and allocate appropriate resources, such as memory and processing power (GPU/CPU). An S3 bucket is designated for storing files and model outputs.

\subsection{Data Preparation}
SageMaker’s Ground Truth tool offers flexible image labeling capabilities, including bounding box annotations for images stored in our S3 bucket. It supports both single and multi-class labeling and allows users to enlist additional workers for annotation tasks. The labeled data, along with a manifest file containing bounding box coordinates and image paths, is stored in the S3 bucket, enabling seamless integration into the training pipeline.

\subsection{Model Deployment}
SageMaker’s Jupyter notebooks are utilized for code development, debugging, and deployment, offering seamless access to files stored in S3. This functionality simplifies data handling and ensures that outputs are stored as specified. Although this study did not explore SageMaker's automatic algorithm selection feature, GPU-based Jupyter notebooks were manually created for the efficient training of various YOLO models.

\subsection{SageMaker Instance Specification}
SageMaker supports a range of instance types tailored for various machine learning workloads. For this project, we utilized the \texttt{ml.p3.2xlarge} and \texttt{ml.p3.8xlarge} instances to support both single-GPU and multi-GPU tasks \cite{sagemakerspec}. The \texttt{ml.p3.2xlarge} instance, featuring one NVIDIA Tesla V100 GPU, 8 vCPUs, and 61 GB of memory, provided an optimal balance of computational power for smaller-scale deep learning tasks and development. In contrast, the \texttt{ml.p3.8xlarge} instance, equipped with 4 GPUs, 32 vCPUs, and 244 GB of memory, was selected for training more complex models that required higher computational capacity. For instance, the \texttt{Yv9\_g-c\_2} model, with its greater number of layers, parameters, and intricate architecture, was trained using two GPUs to maximize training efficiency. All other models were trained using the \texttt{ml.p3.2xlarge} instance. This strategic use of instances significantly expedited training and improved the performance of image recognition and natural language processing tasks compared to CPU-based alternatives.

\section{Experimental Results}
\label{results}
In this section, we present our results in various tables. In Table \ref{tab:modelspec}, we use shortened forms of model names to describe their respective details. In Table~\ref{tab:yolohardware}, we provide a comprehensive overview of various performance metrics and hardware specifications for the YOLO models. We include details such as the number of epochs, the time taken for each epoch to train the model. Additionally, we present information about the number of GPUs utilized during training (which impacts training time), and GFLOPs, offering insights into the computational capacity. Furthermore, we list parameters representing the model's trainable weights and biases, which are essential components influencing its performance and behavior. Finally, a comparison of model performances are presented in Table~\ref{tab:yoloperf}, highlighting precision, recall, and mAP scores.

\begin{table}[h]
    \centering
    \caption{Specifications of the Models}
    \label{tab:modelspec}
    \begin{tabular}{|lcl|}
        \hline
        \textbf{Models} & \textbf{Full Forms} & \textbf{Description}\\
        \hline
        Yv5s         & YOLO version 5 & Small edition \\
        \hline
        Yv5m         & YOLO version 5 & Medium edition \\
        \hline
        Yv8s         & YOLO version 8 & Small edition \\
        \hline
        Yv8m         & YOLO version 8 & Medium edition \\
        \hline
        \multirow{2}{*}{Yv9\_g-c} & \multirow{2}{*}{YOLO version 9} & Generalized Efficient Layer\\ 
        & &  Aggregation Network (GELAN) \cite{wang2024yolov9}\\ 
        \hline
    \end{tabular}
\end{table}

\subsection{Hardware Specific Model Performances}
The hardware performance evaluation of YOLO models in Table~\ref{tab:yolohardware} highlights key metrics, including the number of epochs, training time per epoch, number of GPUs used, GFLOPS, and parameter count. GFLOPS is a measure of computational complexity, representing the number of billions of operations a model can perform per second. From the table, we observe that smaller versions of YOLO models (e.g., Yv5s) have lower GFLOPS and parameter counts, resulting in shorter training times per epoch and requiring fewer computational resources. The training time per epoch for Yv5s is minimal, aligning with its lightweight design and lower parameter count. Conversely, medium models like Yv8m show an increase in both GFLOPS and parameter count, contributing to a moderate training duration while still being feasible to train on a single GPU.

\begin{table}[t]
    \centering
    \caption{Hardware Specific Performance Evaluation of YOLO Models}
    \label{tab:yolohardware}
    \begin{tabular}{|lcrcrr|}
        \hline
        \textbf{Model} & \textbf{Epochs} & \textbf{Time/epoch} & \textbf{GPUs} & \textbf{GFLOPS} & \textbf{Params.} \\
        \hline
        Yv5s         & 146 & 0.00269 & 1 &  15.8 & 7,015,519 \\
        \hline
        Yv5m         & 156 & 0.00293 & 1 &  47.9 & 20,856,975  \\
        \hline
        Yv8s         & 158 & 0.00087 & 1 &  28.4 & 11,126,358  \\
        \hline
        Yv8m         & 180 & 0.00151 & 1 &  79.1 & 25,857,478  \\
        \hline
        Yv9\_g-c\_1     & 267 & 0.00470 & 1 & 102.5 & 25,412,502  \\
        \hline
        Yv9\_g-c\_2     & 209 & 0.00353 & 2 & 102.5 & 25,412,502  \\
        \hline
    \end{tabular}
    \vspace{-.35em}
\end{table}
  
Larger models such as YOLOv9, which have the highest GFLOPS and parameter counts, demonstrate the need for significant computational power. The training time per epoch is longer, especially when using a single GPU. However, training on multiple GPUs (e.g., Yv9\_g-c\_2 trained on 2 GPUs) helps to reduce the training time per epoch, showcasing the scalability of the architecture. Models with higher GFLOPS, such as \texttt{Yv9\_g-c\_1} and \texttt{Yv9\_g-c\_2}, indicate greater computational demand due to more complex structures, which can enhance detection accuracy but require robust hardware and longer training durations. In contrast, models with lower GFLOPS, like Yv5s, are more computationally efficient, allowing faster training with potentially reduced accuracy.

\subsection{Overall Models Performance}
In Table~\ref{tab:yoloperf}, we report the model performances, highlighting Precision, Recall, mAP50, and mAP50-95. Precision indicates the proportion of correctly identified positive cases among all cases flagged as positive, reflecting the model's ability to avoid false positives. Recall measures the proportion of correctly identified positive cases out of all actual positives, demonstrating the model's effectiveness in detecting relevant instances. The Mean Average Precision at a 50\% Intersection over Union (IoU) threshold (mAP50) calculates average precision across classes, while mAP50-95 provides a more comprehensive assessment by evaluating performance across IoU thresholds from 50\% to 95\%.

\begin{table}[h]
    \centering
    \caption{Performance Evaluation of YOLO Models}
    \label{tab:yoloperf}
    \begin{tabular}{|lcccc|}
        \hline
        \textbf{Model} & \textbf{Precision} & \textbf{Recall} & \textbf{mAP50} & \textbf{mAP50-95}\\
        \hline
        Yv5s         & 50.3 & \textbf{55.6} & 51.4 & 20.4 \\
        \hline
        Yv5m         & 45.8 & 53.2 & 48.8 & \textbf{20.9} \\
        \hline
        Yv8s         & 39.3 & 40.4 & 48.9 & 20.8 \\
        \hline
        Yv8m         & 47.1 & 47.8 & \textbf{52.2} & 19.5 \\
        \hline
        Yv9\_g-c\_1 & 37.5 & 57.1 & 47.8 & 18.6 \\
        \hline
        Yv9\_g-c\_2 & \textbf{52.6} & 47.8 & 45.8 & 16.3 \\
        \hline
    \end{tabular}
\end{table}

The results show that the Yv9\_g-c\_2 model (with 2 GPUs) achieved the highest precision score of 52.6\%, followed closely by Yv5s at 50.3\%. Yv5m, Yv8m, and Yv9\_g-c\_1 (with single GPU each) also performed well, with precision scores between 45.8\% and 47.1\%. Yv8s had the lowest precision score at 39.3\%. For recall, Yv5s led with a score of 55.6\%, while Yv9\_g-c\_1 (with a single GPU) demonstrated a score of 57.1\%. The remaining models exhibited recall scores ranging from 40.4\% to 47.8\%.
Yv8m achieved the highest mAP50 score of 52.2\%, closely followed by Yv5s with 51.4\%. Yv5m, Yv8s, and Yv9\_g-c\_1 also performed well with mAP50 scores ranging from 45.8\% to 48.9\%. Yv9\_g-c\_2 exhibited the lowest mAP50 score of 45.8\%. The experiment shows, the highest mAP50-95 score of 20.4\%, indicating its effectiveness across a range of IoU thresholds. Yv8m and Yv8s also performed well with mAP50-95 scores of 19.5\% and 20.8\%, respectively. However, Yv9\_g-c\_2 exhibited the lowest mAP50-95 score of 16.3\%.

\begin{figure}[htbp]
  \centering
  \begin{subfigure}{0.48\linewidth} 
    \includegraphics[width=\linewidth, height=4.25cm]{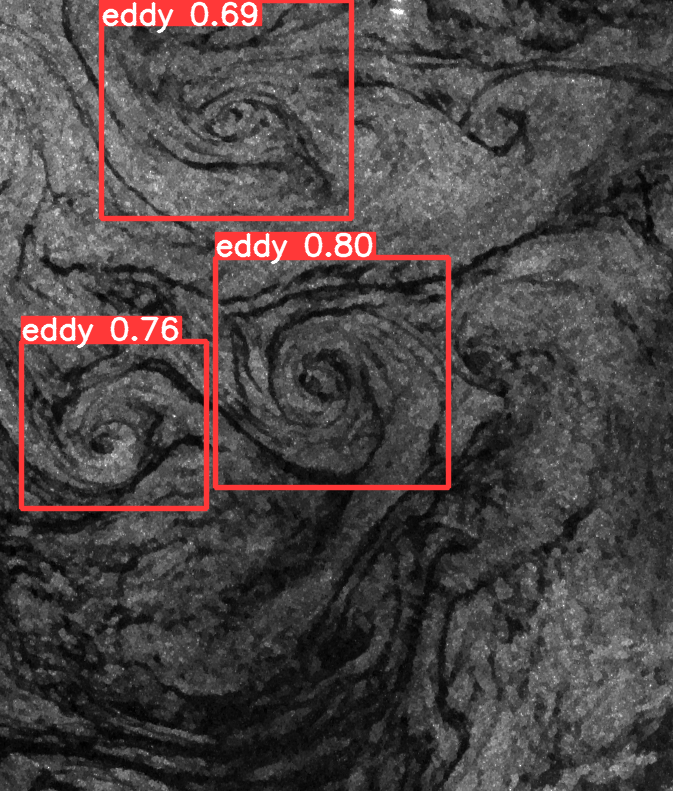}
    \caption{Yv5s}
    \label{subfig:1y5s}
  \end{subfigure}%
  \hspace{0.01\linewidth} 
  \begin{subfigure}{0.48\linewidth} 
    \includegraphics[width=\linewidth, height=4.25cm]{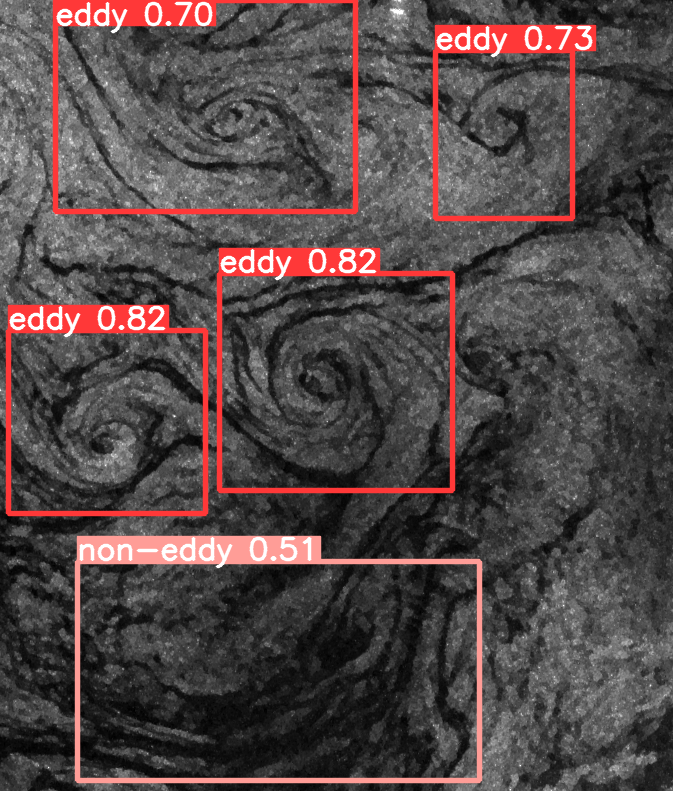}
    \caption{Yv5m}
    \label{subfig:1y5m}
  \end{subfigure}\\[0.2cm]

  \begin{subfigure}{0.48\linewidth} 
    \includegraphics[width=\linewidth, height=4.25cm]{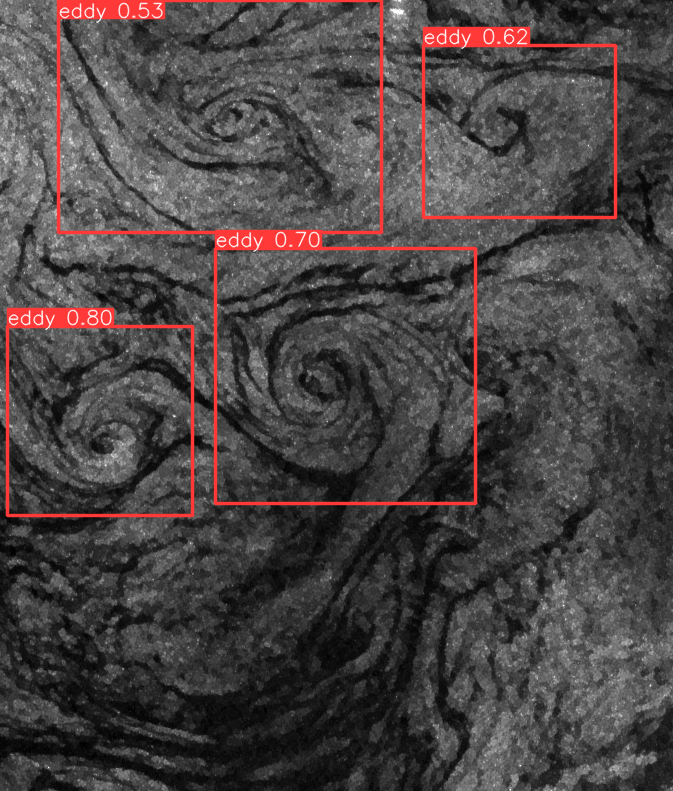}
    \caption{Yv8s}
    \label{subfig:1y8s}
  \end{subfigure}%
  \hspace{0.01\linewidth} 
  \begin{subfigure}{0.48\linewidth} 
    \includegraphics[width=\linewidth, height=4.25cm]{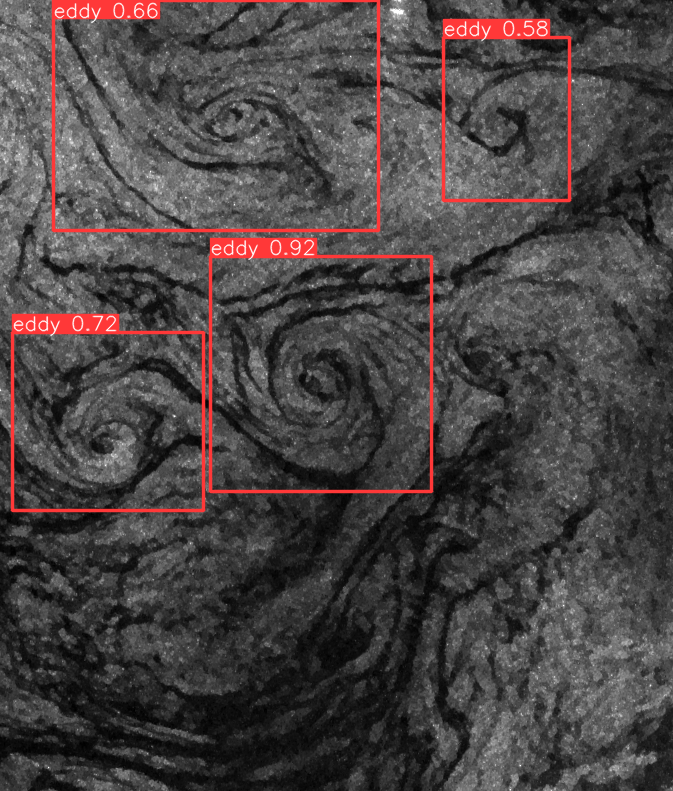}
    \caption{Yv8m}
    \label{subfig:1y8m}
  \end{subfigure}\\[0.2cm]

  \begin{subfigure}{0.48\linewidth} 
    \includegraphics[width=\linewidth, height=4.25cm]{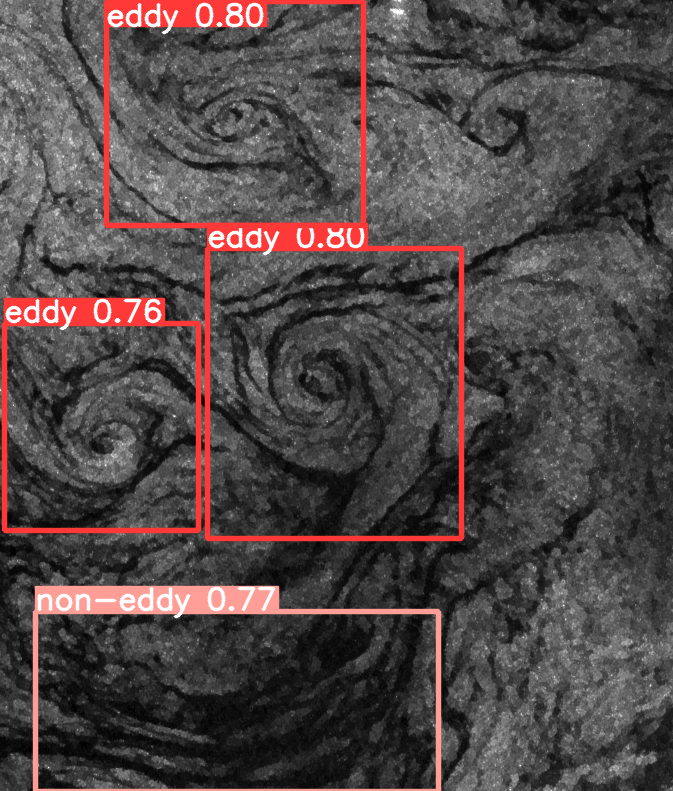}
    \caption{Yv9\_g-c\_1}
    \label{subfig:1y9gc1}
  \end{subfigure}%
  \hspace{0.01\linewidth} 
  \begin{subfigure}{0.48\linewidth} 
    \includegraphics[width=\linewidth, height=4.25cm]{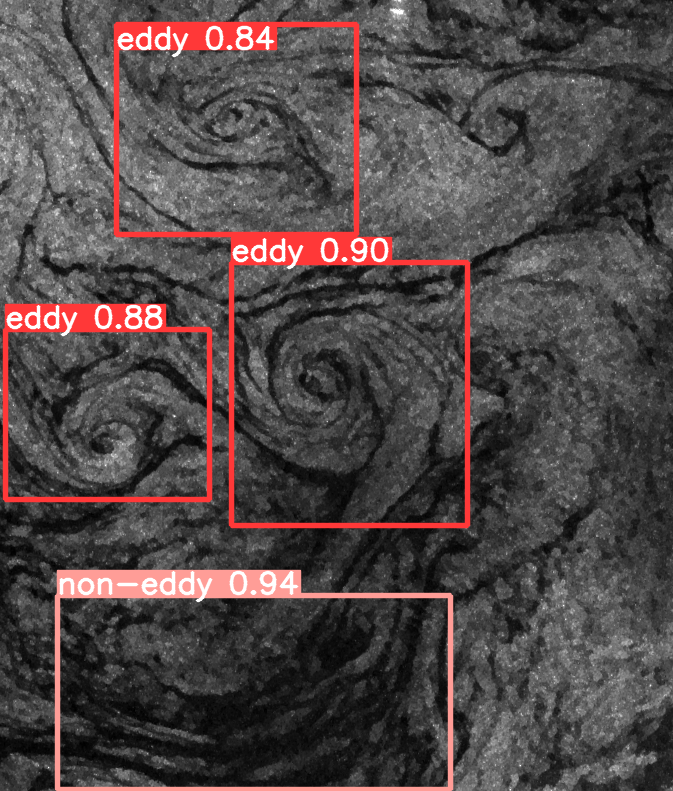}
    \caption{Yv9\_g-c\_2}
    \label{subfig:1y9gc2}
  \end{subfigure}

  \caption{Localization of Ocean Eddies: A Comparative Analysis of Various Models Applied to an Image.}
  \label{fig:loc1}
\end{figure}

\subsection{Comparison of Localization of Objects}
\begin{figure}
  \centering
  \begin{subfigure}{0.48\linewidth}
    \includegraphics[width=\linewidth, height=4.25cm]{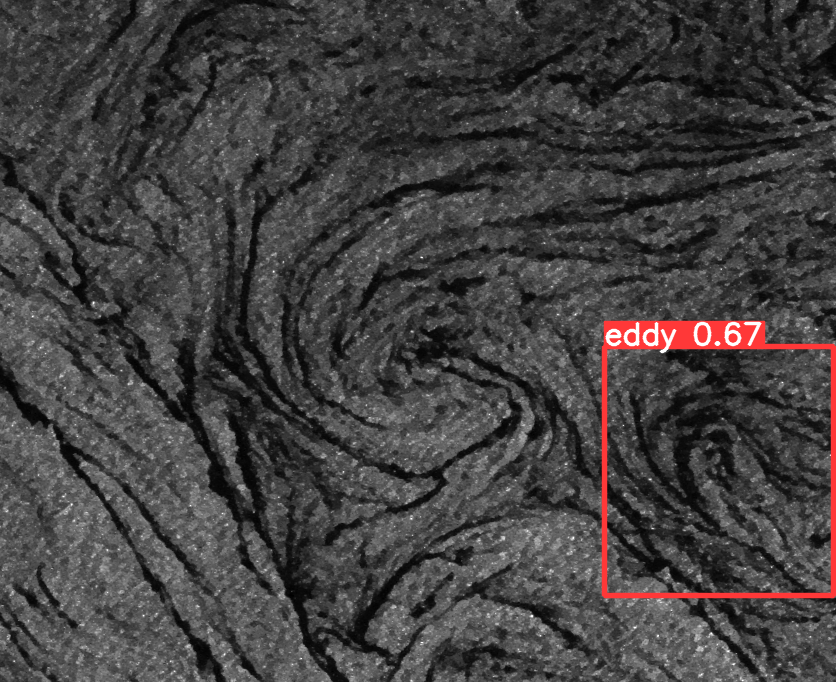}
    \caption{Yv5s}
    \label{2y5s}
  \end{subfigure}%
  \hspace{0.01\linewidth} 
  \begin{subfigure}{0.48\linewidth}
    \includegraphics[width=\linewidth, height=4.25cm]{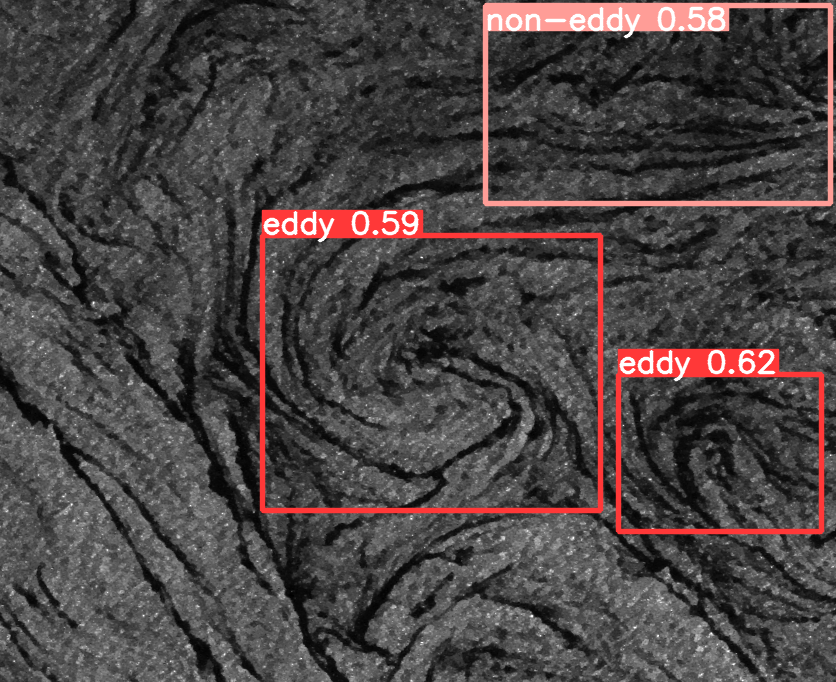}
    \caption{Yv5m}
    \label{2y5m}
  \end{subfigure}\\[0.2cm]

  \begin{subfigure}{0.48\linewidth}
    \includegraphics[width=\linewidth, height=4.25cm]{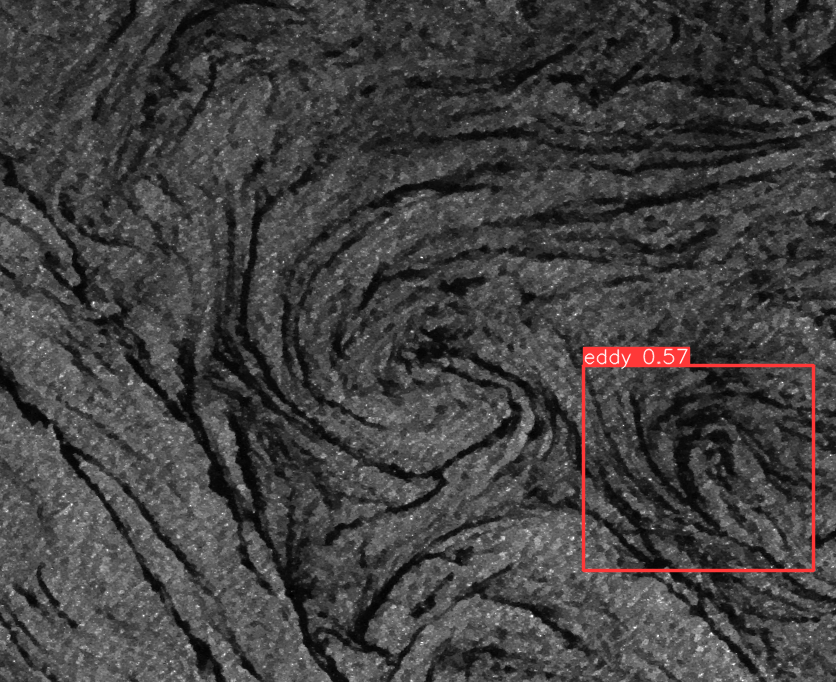}
    \caption{Yv8s}
    \label{2y8s}
  \end{subfigure}%
  \hspace{0.01\linewidth} 
  \begin{subfigure}{0.48\linewidth}
    \includegraphics[width=\linewidth, height=4.25cm]{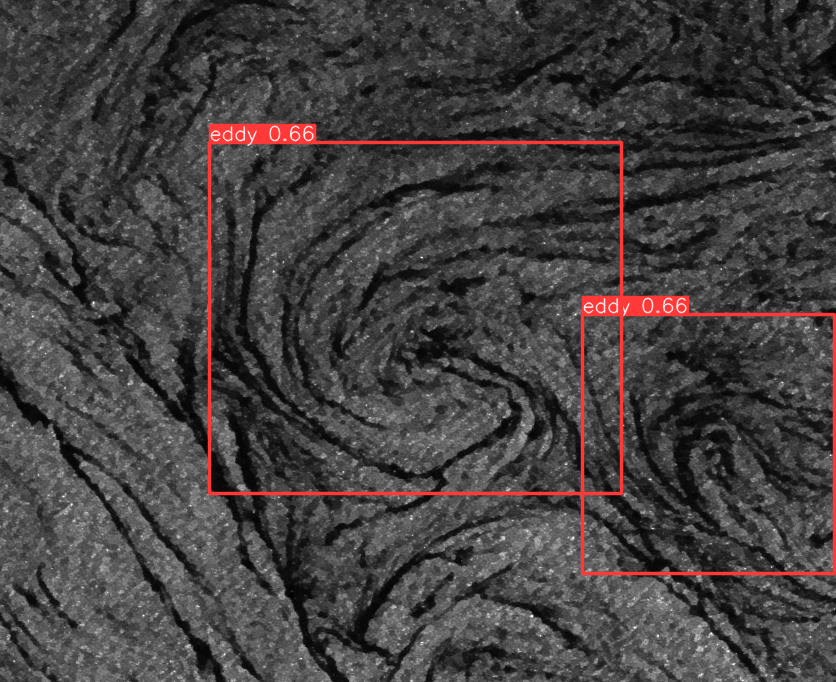}
    \caption{Yv8m}
    \label{2y8m}
  \end{subfigure}\\[0.2cm]

  \begin{subfigure}{0.48\linewidth}
    \includegraphics[width=\linewidth, height=4.25cm]{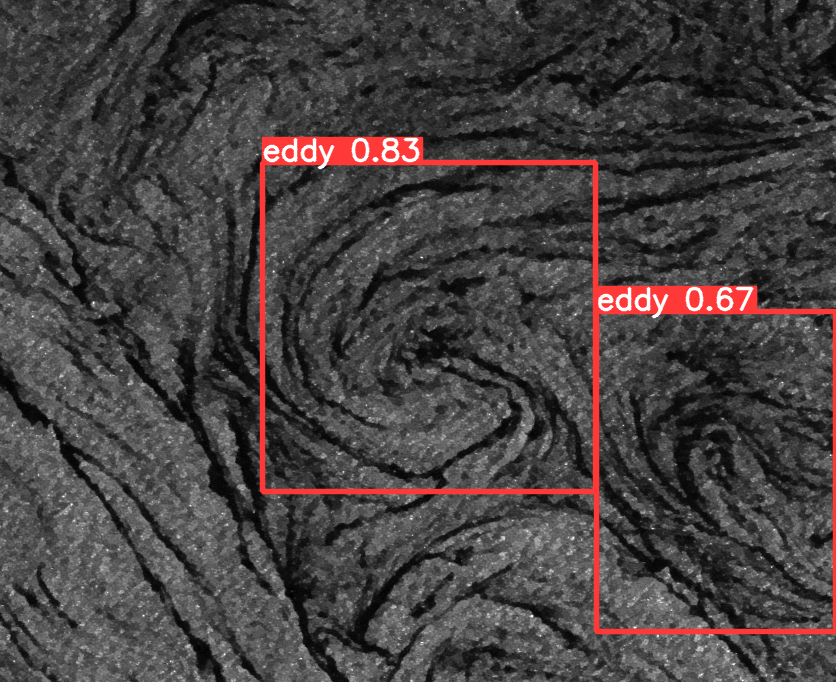}
    \caption{Yv9\_g-c\_1}
    \label{2y9gc1}
  \end{subfigure}%
  \hspace{0.01\linewidth} 
  \begin{subfigure}{0.48\linewidth}
    \includegraphics[width=\linewidth, height=4.25cm]{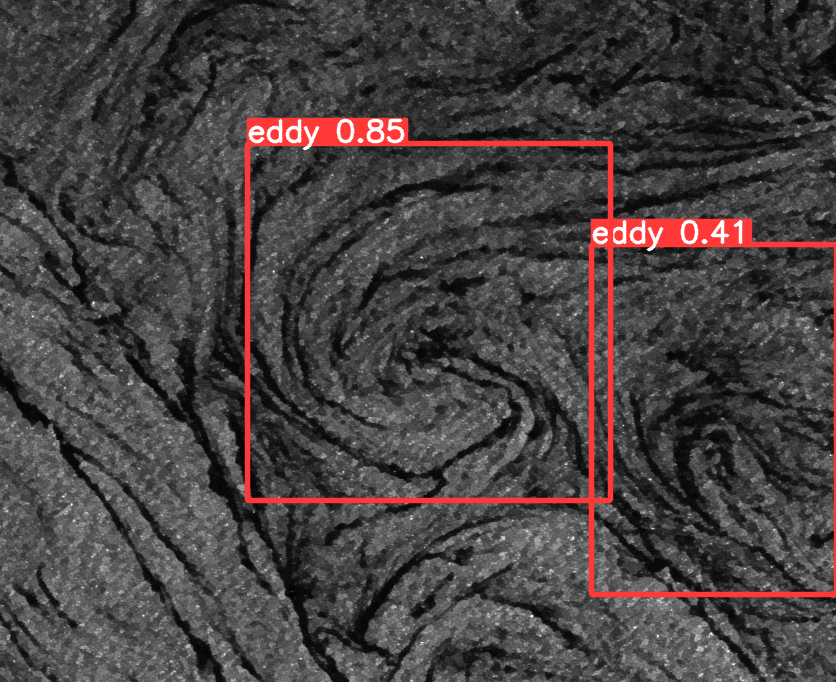}
    \caption{Yv9\_g-c\_2}
    \label{2y9gc2}
  \end{subfigure}

  \caption{Localization of Ocean Eddies: Comparative Evaluation Across Multiple Models Using Additional Image.}
  \label{fig:loc2}
\end{figure}

In addition to evaluating the models' performance, we conducted a comparison of their localization capabilities, as depicted in Figures \ref{fig:loc1} and \ref{fig:loc2}. Each sub-figure in both plots represents the localization performance of different YOLO subcategories within each model. However, YOLOv9 follows a different naming convention. During the experiments, we were limited to using only the YOLOv9\_g-c, as it was the only version available at the time of the experiments were conducted.

In Figure \ref{fig:loc1}, we observe that all models are capable of detecting most of the eddies within the images, albeit with varying confidence scores across different edition (in Table~\ref{tab:modelspec}) for each model. For instance, Sub-figures \ref{subfig:1y5s} and \ref{subfig:1y5m} pertain to YOLOv5, where the medium version detects more eddies compared to the smaller version and also captures a "non-eddy" that was missed by the smaller version. Similarly, for YOLOv8 smaller and medium versions, a similar pattern is observed, albeit without the detection of the "non-eddy" in the smaller version. Conversely, YOLOv9 (both \texttt{Yv9\_g-c\_1} and \texttt{Yv9\_g-c\_2}) detected all cases with higher confidence scores.

Furthermore, we conducted a comparison of all models in Figure \ref{fig:loc2}, where we observed that the medium category of both YOLOv5 (sub-figure \ref{2y5m}) and YOLOv8 (sub-figure \ref{2y8m}) performed better compared to the small category. Comparing with YOLOv9 (both \texttt{Yv9\_g-c\_1} and \texttt{Yv9\_g-c\_2}), it outperformed all models in terms of detection and increased confidence scores.

\begin{figure}
  \centering
  \begin{minipage}{0.485\linewidth} 
    \includegraphics[width=\linewidth, height=4.25cm]{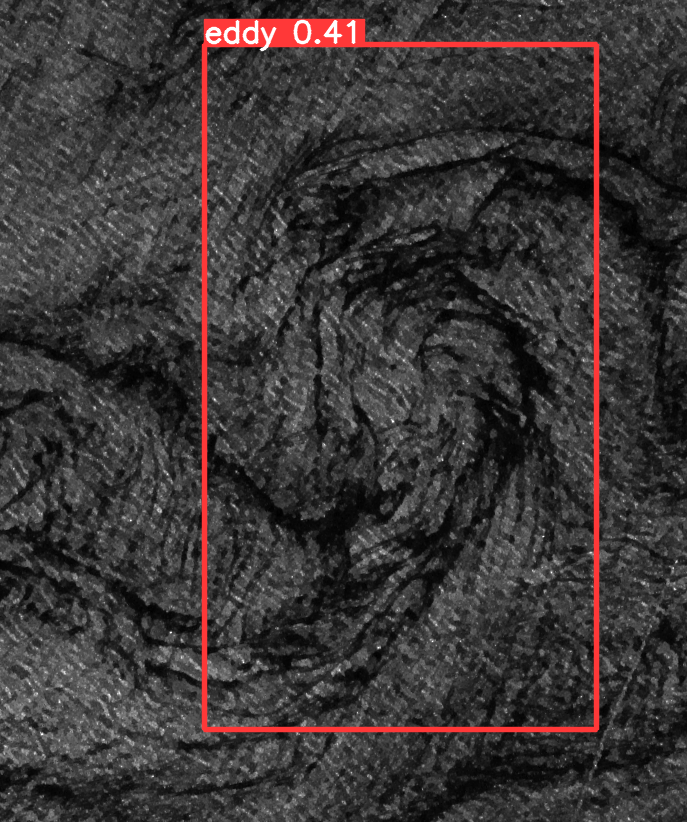}
    \subcaption{}
    \label{subfig:4y9gc1}
  \end{minipage}%
  \hspace{0.01\linewidth} 
  \begin{minipage}{0.485\linewidth} 
    \includegraphics[width=\linewidth, height=4.25cm]{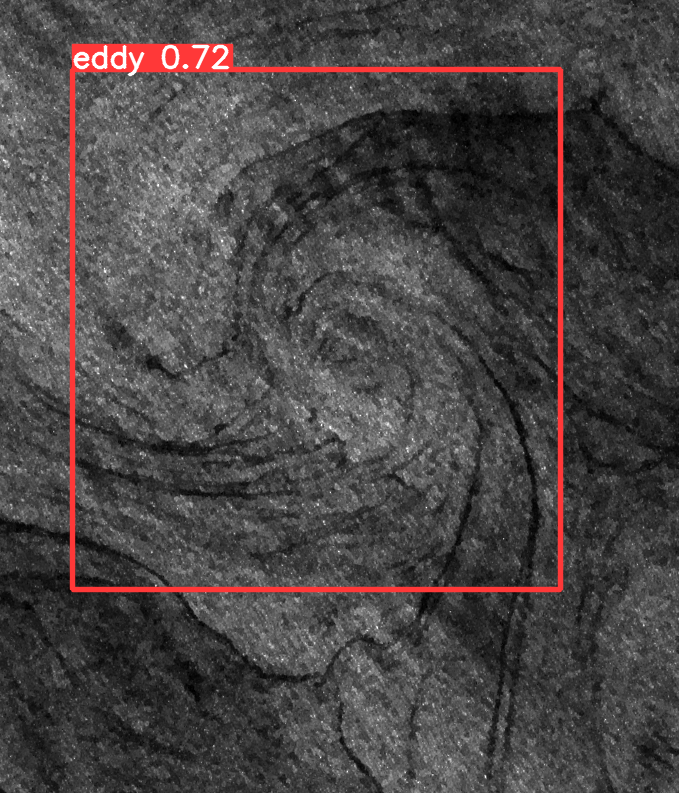}
    \subcaption{}
    \label{subfig:5y9gc1}
  \end{minipage}\\[0.2cm]

  \begin{minipage}{0.485\linewidth} 
    \includegraphics[width=\linewidth, height=4.25cm]{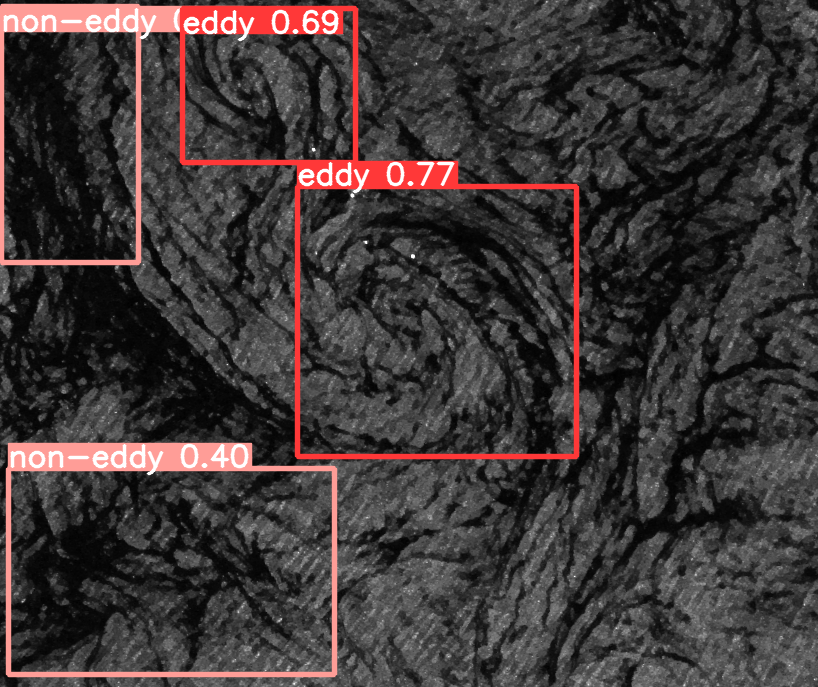}
    \subcaption{}
    \label{subfig:6y9gc1}
  \end{minipage}%
  \hspace{0.01\linewidth} 
  \begin{minipage}{0.485\linewidth} 
    \includegraphics[width=\linewidth, height=4.25cm]{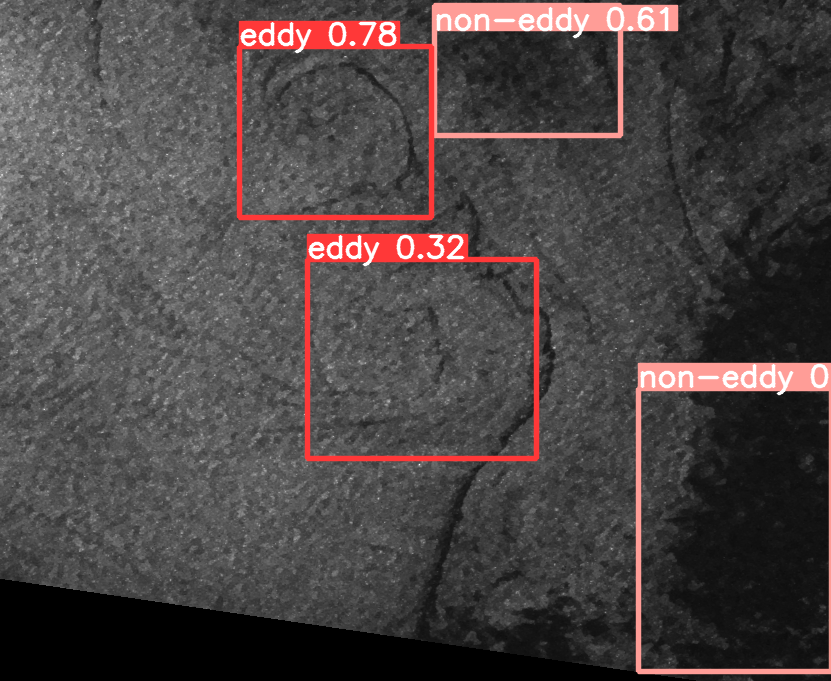}
    \subcaption{}
    \label{subfig:7y9gc1}
  \end{minipage}

  \caption{Spotlighting Ocean Eddies: Exclusive Detection via YOLOv9 Model only.}
  \label{fig:loc3}
\end{figure}

Finally, we compared another set of images as depicted in Figure~\ref{fig:loc3}. In this figure, all detections were made using YOLOv9 (both  \texttt{Yv9\_g-c\_1} and \texttt{Yv9\_g-c\_2}) exclusively, as both YOLOv5 and YOLOv8 failed to detect any of the eddies within those images. This clearly demonstrates the success of YOLOv9 in this scenario. These images have a slightly complex structure, yet YOLOv9 was able to locate them almost accurately.

The comparison study highlights the strengths and weaknesses of each model across various performance metrics and hardware specifications. YOLOv5s and YOLOv8m stand out for their competitive precision, recall, and mAP50, along with efficient training times. However, the optimal model choice depends on task-specific requirements and constraints. Notably, YOLOv9 outperformed YOLOv5 and YOLOv8 in localizing ocean eddies (Figure~\ref{fig:loc3}), even under challenging conditions. Despite this, YOLOv5 achieved the highest precision and recall. This experiment used a smaller dataset with larger individual file sizes, and a larger dataset would provide a more comprehensive performance assessment.

\section{Discussions}
\label{disc}

\begin{table*}[h]
    \centering
    \caption{SageMaker in a Nutshell}\label{tab:sagemaker}
    \begin{tabular}{|m{4cm} | m{13cm}|}
        \hline 
        \multicolumn{1}{|c|}{\textbf{Feature}} & \multicolumn{1}{c|}{\textbf{Description}} \\ 
        \hline
        Usability & 
        \begin{itemize} 
            \item Need to configure and install libraries prior to use.
            \item CPU or GPU access to notebooks.
            \item Both the notebook and scripts are executable.
            \item Notebook instances require turning on and off for usage.
            \item Documents need to be saved before turning off.
            \item SageMaker does not have enough memory to store files except for smaller datasets; S3 bucket is used to store large datasets.
            \vspace{-\baselineskip}
        \end{itemize} \\ 
        \hline
         Machine Learning as a Service & 
        \begin{itemize}
            \item SageMaker is designed for seamless MLaaS experience with ground truth, data labeling, clustering, algorithm and model selection options. However, the whole pipeline deployment is cumbersome by selecting available models.
            \item External API call is allowed through SageMaker, enabling the use of MLaaS.
            \item Notebooks running in an idle state have a price too.
            \vspace{-\baselineskip}
        \end{itemize} \\ 
            \hline
        Performance & 
        \begin{itemize} 
            \item SageMaker provides a fully configured environment and computing power with a combination of either GPU or CPU, faster read-write capability of the SSD storage.
            \item Integration of Lambda services, RESTful API call, and SageMaker is a great combination for ML service deployment.
            \vspace{-\baselineskip}
            \end{itemize} \\
            \hline
        Cloud-based Services for Earth Observation Research & 
        \begin{itemize} 
            \item Data accessibility, labeling, training, and storing using SageMaker is a good fit to deploy ML models, which we presented through the Ocean Eddy project.
            \item For continuous data and result integration, model deployment and sharing between collaborators is convenient. 
            \item SageMaker models are not shareable directly; however, they can be accessed with permitted users in a group or with secret and public key access.
            \item RESTful API integration is allowed in SageMaker to execute notebooks and access files in S3.
            \vspace{-\baselineskip}
            \end{itemize}\\
          \hline
    \end{tabular}
\end{table*}

This section focuses into the challenges we faced when deploying models using SageMaker and outlines potential directions for future work. One of the primary goals of this study was to evaluate the ease and flexibility of using SageMaker for Earth science research. We aimed to explore its built-in features comprehensively and assess its user-friendliness. Given the unique challenges associated with satellite data, including large file sizes and varying formats, it is crucial for a platform to offer robust support tailored to these needs. In the following subsections, we detail some of our key findings regarding SageMaker's shortcomings. We also propose future research avenues for Earth science projects that would benefit from enhanced capabilities and support provided by cloud platforms, ensuring they meet the demands of complex data processing and analysis.

\subsection{Limitations}
Despite its notable features, Amazon SageMaker presented several significant shortcomings during our experimentation as follows.

\textbf{\textit{I) Labeling Limitations.}} A major limitation arose from the restriction on the number of images that can be labeled in a single task, capped at ten images at most. This constraint poses a considerable obstacle when training models with larger datasets.

\textbf{\textit{II) Worker Assignment Challenges.}} The mandatory assignment of workers for labeling tasks posed another challenge. Without an assigned worker, the labeling process cannot proceed. Moreover, simultaneous labeling by multiple workers is not supported, leading to failed processes if multiple workers attempt to label images concurrently. This limitation significantly hampers efficiency, as the functionality fails to fulfill its intended purpose.

\textbf{\textit{III) Model Selection Issues.}} Despite documentation indicating that users can select pre-configured models for training, we encountered challenges in implementing this feature, even with the paid version of SageMaker. As a result, we opted to deploy our own YOLO models to complete the tasks.

\textbf{\textit{IV) Image Format Issues.}} SageMaker encountered difficulty in displaying Ocean eddy images converted from GEO-TIFF to PNG using OpenCV or Matplotlib. This obstacle was successfully overcome by utilizing the GDAL library. Nevertheless, this solution also imposes constraints on the accessibility and usability of SageMaker in broader contexts.

\subsection{Potential Future Research Opportunities} 
In the realm of eddy detection, we see several avenues for further research specific to enhancing the detection eddies in satellite imagery as follows.

\textbf{\textit{I) Evaluating Eddy Diameter.}} Exploring the diameter of eddies would enhance our ability to measure them accurately. This investigation could shed light on whether eddies undergo changes in size or shape due to climate variations, providing valuable insights into environmental dynamics.

\textbf{\textit{II) Precise Eddy Center Localization.}} Refining techniques to pinpoint the center of eddies would enable comprehensive eddy detection, even in cases where parts of the eddies are obscured in images. Accurately locating the center facilitates more precise diameter measurements.

\textbf{\textit{III) Multiple Satellite Co-registration.}} Ocean eddies can often be identified from different satellite measurements, such as level-2 sea surface temperature from VIIRS at 750 m grid posting. The recently-launched Surface Water and Ocean Topography mission also provide a great opportunity to study these small-scale eddies. When the two satellites scan the same eddy simultaneously, one often can improve the eddy identification as well as deriving higher-order quantities that is related to climate processes, such as upper ocean heat uptake and carbon uptake.

\textbf{\textit{IV) Model Validation with Diverse Earth Science Data.}} Our experiment utilized a limited dataset. To thoroughly assess model efficacy, it's imperative to test them with diverse categories of earth science data. Such comprehensive testing not only reveals the true performance of the models but also maximizes the potential of cloud computing through parallel architecture and additional GPUs.

\textbf{\textit{V) Enhancing SageMaker Usability.}} Despite its intended purpose to facilitate data enthusiasts, our experiments uncovered significant limitations in SageMaker. Addressing these limitations by the platform provider could unlock more opportunities for data enthusiasts in the future.

We further highlight the key aspects of SageMaker in Table~\ref{tab:sagemaker} in terms of deploying AI models. Overall, it is an essential tool that provides state-of-the-art features for the research community. It is expected that SageMaker will continue to enhance its capabilities by expanding the availability of libraries and incorporating features for easy accessibility, such as data conversion, data splitting, and streamlined access to models.

\section{Conclusions}
\label{conc}
Our study leveraged Amazon SageMaker, a cloud-based service, for Ocean Eddy localization, which has potential applications in Earth observation. We employed various versions of the YOLO model, including versions 5, 8, and 9, to localize Ocean Eddies. The results of our experiments fulfilled the objective of this work, which was to measure and compare the performance of different YOLO models on the cloud platform.

However, our experience revealed significant challenges associated with deploying models using SageMaker, particularly when utilizing the ground truth feature for AI-based model deployment. These challenges highlight the need for further refinement and improvement in the SageMaker platform to streamline the deployment process, especially when handling big data problems in such cloud environments.

Looking ahead, we are eager to address the potential research opportunities identified during our study. This includes exploring methodologies to enhance the efficiency and usability of SageMaker for deploying AI models, as well as investigating new avenues for object localization and detection in future Earth science projects.


\bibliographystyle{IEEEtran}


\end{document}